\documentclass[11pt,a4paper]{article}
\usepackage[hyperref]{naaclhlt2019}
\usepackage{times}
\usepackage{latexsym}

\usepackage{url}

\usepackage{graphicx}
\usepackage{booktabs}
\usepackage{adjustbox}
\usepackage{verbatim}
\usepackage{amsmath}
\usepackage{multicol}
\usepackage{stmaryrd}

\aclfinalcopy 

\title{Neural Models of the Psychosemantics of `Most'}

\author{Lewis O'Sullivan \\
  Brain and Cognitive Sciences \\
  Universiteit van Amsterdam \\
  {\tt \small lewis.osullivan@student.uva.nl} \\\And
  Shane Steinert-Threlkeld \\
  Institute for Logic, Language and Computation \\
  Universiteit van Amsterdam \\
  {\tt \small S.N.M.Steinert-Threlkeld@uva.nl} \\}

\date{}

\begin{document}

\maketitle

\begin{abstract}
How are the meanings of linguistic expressions related to their use in concrete cognitive tasks? Visual identification tasks show human speakers can exhibit considerable variation in their understanding, representation and verification of certain quantifiers. 
This paper initiates an investigation into neural models of these psycho-semantic tasks.  We trained two types of network -- a convolutional neural network (CNN) model and a recurrent model of visual attention (RAM) -- on the ``most'' verification task from \citet{Pietroski2009}, manipulating the visual scene and novel notions of task duration.  Our results qualitatively mirror certain features of human performance (such as sensitivity to the ratio of set sizes, indicating a reliance on approximate number) while differing in interesting ways (such as exhibiting a subtly different pattern for the effect of image type).  We conclude by discussing the prospects for using neural models as cognitive models of this and other psychosemantic tasks.
\end{abstract}

\section{Introduction}

Semantics -- the scientific study of meaning -- has traditionally studied the truth-conditions of sentences and how the meanings of sub-sentential expressions combine to generate them.  How exactly truth-conditions are represented and then deployed in concrete acts of production and comprehension has often not been seen as belonging to the purview of semantics properly.

A recent line of work, however, has argued that the mental representation of the meanings of expressions bias behavior in cognitive tasks in ways that allow us to adjudicate between truth-conditionally equivalent but representationally distinct semantic theories.  In particular, \citet{Pietroski2009} considered the verification of the sentence ``Most of the dots are yellow''.  The meaning of `most' can be expressed in distinct, but truth-conditionally equivalent ways.  For instance (where, in the running example, $A$ is the set of dots, and $B$ the set of yellow things):
\begin{itemize}
    \item $\llbracket \text{most} \rrbracket(A)(B) = 1 \text{ iff } |A \cap B| > |A \setminus B|$
    \item $\llbracket \text{most} \rrbracket(A)(B) = 1 \text{ iff } \text{ there is } f : A \setminus B \to A \cap B \text{ that is one-to-one, but not onto}$
\end{itemize}
The former says that the number of dots which are yellow is larger than the number of non-yellow dots, while the latter says that the former can be paired off with the latter, with some yellow dots remaining.  Whilst these representations are truth conditionally equivalent, each is associated with a distinct \emph{verification strategy} to evaluate those truth conditions. When deciding whether most of the dots are yellow: the former representation is associated with an algorithm for computing and comparing two cardinalities, while the latter representation is associated with an algorithm for checking whether a certain correspondence between yellow and non-yellow dots exists. Whilst a speaker may be capable of implementing many possible strategies, \citet{Pietroski2009}'s claim is that, all other things being equal, speakers are biased towards using the default strategy associated with their representation.

\citet{Pietroski2009} sought to determine whether speakers prefer one of the above representations by testing which verification strategy they typically use. By manipulating the arrangement of the dots in images against which `most' was verified, they created conditions which should ease the implementation of one of the strategies (e.g. dots arranged in pairs should favour correspondence). 
They found no difference in verification accuracy between three of the four image types used. Participants were significantly more accurate on the remaining image type, which consisted of two paired columns of colour sorted dots. Their analysis suggested that the participants used the columns' lengths as a proxy for set cardinality, rather than using a correspondence strategy. The results of the remaining three image types were explaind very well by a psychophyiscal model of approximate number. Given that this system cannot be used to implement a correspondence strategy, they concluded that the meaning of `most' is best represented in the former way.\footnote{See \citet{Lidz2011} for further research in this direction, distinguishing between more candidate representations.}

In this paper, we begin to develop robust mechanistic \emph{cognitive models} of their sentence verification task to help elucidate the factors underlying the psychosemantics of `most'.  In particular, we are interested in the following question: do various neural models show the potential to be developed into good cognitive models of the meaning of `most'?  A good cognitive model does at least two things: (i) fits human data well and (ii) has movable parameters that enable new predictions to be made.
To address this question, we subjected two different classes of models -- convolutional networks and recurrent models of visual attention -- to the experimental design from \citet{Pietroski2009}, together with an additional and novel manipulation for `task duration' (inspired by \citet{Register2018}).  This allows us to assess the models along both dimensions (i) and (ii).  Our key contributions are:
\begin{itemize}
    \item Subjecting neural models to prominent tasks from the psychosemantics literature.
    \item Operationalizing `task duration' in two distinct ways: depth of a convolutional network, and the number of glimpses in a model of visual attention.
\end{itemize}
The key findings from our experiments are:
\begin{itemize}
    \item Both models exhibit patterns of behavior qualitatively similar to humans, including sensitivity to dot ratio.
    \item The psychophysical model of approximate number fits model data well, with parameters not too far from human participants.
    \item Model performance is effected by the image type in a subtly different way than human performance.
    \item The effect of task duration is more robust for the convolutional networks than for visual attention.
\end{itemize}

After discussing related work in the next section, we outline the hypotheses of our experiment, before a full explanation of our methods and results.  We conclude by discussing the results and outlining future work.

\section{Related Work}

\subsection{``Most'' and the Visual Identification Task} 


As discussed in the introduction, different representations of a quantifier's meaning may reflect different default verification strategies.  This raises the question: given the many psychologically plausible verification strategies, can we determine whether any are favoured by speakers? \citet{Pietroski2009} addressed this question using the methods described above.
Consequently, by identifying where speakers were most accurate, they were able to determine which strategy speakers favour and, thus, how `most' is represented. 

Their results suggested that speakers favour a cardinality comparison strategy, computed via the approximate number system (ANS) \cite{Dehaene1997}. The ANS is a cognitive system for representing magnitudes. Instead of relying on discrete symbols, such as precise cardinalities, the ANS's representations are imprecise and distributed. They can be described using a series of overlapping Gaussian curves across a continuous `number line': each curve's mean is the cardinality which it corresponds to and the standard deviations increase linearly with the cardinality. Thus, the greater the magnitude of a cardinality, the less precise is its ANS representation. Because the ANS follows what's known as Weber's law \cite{Feigenson2004}, the discriminability of any two ANS representations is determined by the extent of their overlap. Consequently, the difficultly of a cardinality comparison made using the ANS is dependent upon the ratio of the cardinalities. For instance, 6:12 is equally as discriminable as 12:24, or 30:60 or 1:2. This is because the distributions of the ANS representations used to describe these ratios overlap by an equal amount --- they each have a Weber ratio of 2.  The dependence of accuracy on ratio follows a psychophysical model that generates what are called Weber curves (to be described precisely in our Results section).

\citet{Pietroski2009} found that these curves fit participant data very well (in three of four image types) and thus suggest that speakers may employ the ANS as a ``numeralising waystation'' to interface with precise cardinal values. This would allow speakers to understand `most' as a cardinality comparison, but to implement it using the imprecise representations of the ANS. Thus, they claim the semantics for ``most'' is specified in a way that includes cardinality comparison.  We will subject neural models to the same experiment, to see whether they exhibit the same reliance on cardinality and approximation behavior.

\citet{Register2018} argued that it is likely that the participants in \citet{Pietroski2009} were implementing a speed-accuracy trade off due to the number (360) and duration (200ms) of the trials. As such, rather than the preferred semantics for ``most'', they suggest ANS usage may be a result of task-based strategising: participants relied on the speed and the low cognitive effort of an ANS-based strategy in order to cope with unrealistically high demands resulting from the brevity and quantity of the trials. They tested this by running several variations of the experiments from \citet{Pietroski2009}. One experiment asked participants to verify a single trial with unconstrained response time (RT).
Participants' RT and accuracy were negatively correlated, as would be expected were they implementing a speed accuracy trade-off. Nonetheless, their self-reports indicated that most participants used a cardinality comparison based strategy (i.e. either counting or estimating). A second experiment also manipulated the number of trials. Participants who completed more trials were more likely to report using an estimation-based strategy. Additionally, participants' RTs for individual trials decreased as they completed trials. Both of these findings suggest that use of the ANS in \citet{Pietroski2009} was in fact due to task-based strategising. These two findings show that while cardinality comparison is the preferred strategy, it may be computed by different means, depending upon the particular context. As such, the semantics of ``most'' are, to a degree, context dependent.\footnote{In a similar vein, \citet{SteinertThrelkeld2015} show that `most' and `more than half' are differently effected by context under working memory load.}  Our model(s) will incorporate an element of this context-sensitivity, by manipulating a variable not yet tested on humans: task duration, i.e. how long each trial takes.

\subsection{Quantifiers and Neural Networks} \citet{SteinertThrelkeld2018} investigated the hypothesis that semantic universals for quantifiers arise because expressions that satisfy a universal are easier to learn than those that do not. 
By treating the verification and falsification of quantified sentences as a sequence classification task, they trained long short-term memory networks (LSTMs) to learn the meaning of various quantifiers. These quantifiers corresponded to one of three universals (quantity, monotonicity or conservativity), and came in pairs: a real one satisfying the universal and a hypothetical one that does not. By observing whether the LSTMs could learn the expressions satisfying the universals faster (and by extension, more easily), they were able to test this hypothesis. They found that the LSTMs were able to learn to verify expressions which satisfied the quantity and monotonicity universals faster than those which did not, confirming their hypothesis. Not only does this show neural networks are capable of verifying quantifiers, but it suggests that they may do so in a similar way to human speakers.  Nevertheless, their motivation was of a more abstract and theoretical nature; consequently, the networks are not tested on a concrete psycholinguistic task and compared to human performance, as we do here.

\citet{Kuhnle2018} aimed to show how psycholinguistic tasks may provide more informative methods for evaluating how neural networks solve natural language tasks. They trained the FiLM visual question-answering model from \citet{Perez2018a} (a CNN + GRU hybrid) to complete a version of the VIT. Using the Shape World framework \cite{Kuhnle2017}, they generated stimuli consisting of images containing coloured shape objects, a corresponding quantifier statement and a truth value for that statement. The objects were either entirely one colour but from two different shape sets (e.g. red squares and circles) or vice versa (e.g. red and blue squares). The ratio and arrangement of the objects was manipulated. The object set ratios ranged linearly from 1:2 to 7:8, and no image contained more than 15 objects. The objects were either randomly distributed, sorted into contrasting pairs which were randomly distributed, or partitioned by contrasting feature. They trained two instances of the network. The ``Q-half'' network trained on stimuli with ``less/more than half'' statements, whereas the ``Q-full'' network trained on stimuli with a broader range of quantifier statements (e.g. ``some'' and modified numerals such as ``at least 4''). Both networks' test phases exclusively used ``less/more than half'' statements.

Although they found differences in performance according to object arrangement, these did not indicate that the networks favoured any one verification strategy. They suggest the networks may have learned an ``adaptive strategy'' to optimise performance across trial types. Both networks attained high accuracy (100-72\% between ratios 1:2 and 7:8) and became less accurate as the object set ratios became more balanced. The Q-full network was also tested on an evaluation set including the object ratios 8:9, 9:10 and 10:11 (and consequently 17-21 objects). By fitting Weber curves to these data, they found the network's Weber fraction was similar to human speakers'. They interpreted these last two findings as evidence that the network learned an ANS-like system.  While these are promising results, because of different motivations, their stimuli differ in certain ways from those used by \citet{Pietroski2009}, which prevents their models from being cognitive models of the latter task.  Moreover, they have no operationalisation of task duration, to see in what way that affects performance. 

\section{Hypotheses}

In the present experiment, we trained two types of neural network to complete a close replica of the VIT in \citet{Pietroski2009}, with one major addition: we also manipulate task duration (as operationalised by parameters of our neural networks). Based on the VIT research with human speakers discussed in the previous section, we selected three `behavioural traces' which neural networks ought to exhibit if they verify ``most'' in an algorithmically similar manner to human speakers. As such, replicating these behavioural traces is essential for the models to be candidate cognitive models.  Note that we do not assume these traces correspond to, or are necessary evidence of underlying algorithmic similarity between neural networks and human speakers.  However, such similarities would be sufficient causes of these traces. The behavioural traces and their associated hypotheses are:
\begin{enumerate}
    \item ANS usage: Network accuracy is negatively correlated with the stimulus’ dot ratio size.
    \item Verification strategy preference: Network accuracy is dependent upon the arrangement of the stimulus.
    \item Speed-accuracy trade-off: Network accuracy is positively correlated with an appropriate operationalisation of task duration.
\end{enumerate}

\begin{figure*}[ht]
\begin{tabular}{cccc}
\includegraphics[width=0.22\textwidth]{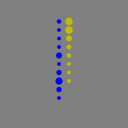}
&
\includegraphics[width=0.22\textwidth]{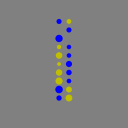}
&
\includegraphics[width=0.22\textwidth]{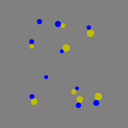}
&
\includegraphics[width=0.22\textwidth]{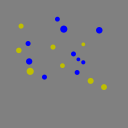}
\end{tabular}
\caption{Example stimuli. All four have a ratio of 5:4, have a positive truth value (i.e. most of the dots are blue). From left to right, image types are: column pairs sorted, column pairs mixed, scattered pairs and scattered random.}
\label{fig:stimuli}
\end{figure*}

\section{Methods}

We generated a range of dot matrix stimuli, each of which consisted two dot sets in a particular ratio and spatial arrangement. Like \citet{Pietroski2009}, we used up to 22 total dots per image, in ratios from 1:2 to 9:10 in one of four arrangements. These were: column pairs sorted (parallel columns of colour sorted dots), column pairs mixed (unsorted parallel columns), scattered pairs (randomly distributed colour-contrast dot pairs) and scattered random (randomly distributed dots). Figure~\ref{fig:stimuli} contains an example of each. Each image was labelled with a truth value for the statement ``Most of the dots are blue''. The stimuli were split into a training set (18000 images), a validation set, and a test set (3600 images each). All three sets were balanced to contain equal proportions of each ratio/image type/truth-value combination.  While we refer to our dot sets as blue and yellow for consistency with the existing literature, we made the input to the networks grayscale in order to reduce dimensionality. 

We used two types of neural network and as such, ran two adjacent experiments. The first of these was an off-the-shelf convolutional neural network (CNN) architecture: the VGG networks from \citet{Simonyan2014}. 

The second was a variation of the recurrent model of visual attention (RAM) from \citet{Mnih2014}.\footnote{In particular, the glimpse network described below did not have convolutional layers and used vector addition instead of component-wise multiplication in \citet{Mnih2014}.} This model processes its input serially in a manner that aims to replicate the saccades and fixations of human visual attention. It does this by taking a series of retina-like samples (called `glimpses') of its `environment' in order to extract the information needed to determine the best location for future glimpses and to solve its task.  This process of visual search and attention reflects a core component of human visual scene representation \cite{Rensink2000, Hayhoe2005, Wolfe2017}.

The network processes an image by using several `sub-networks' operating across a number of time steps ($t$), as depicted in Figure~\ref{fig:ram}:
\begin{figure}[ht]
    \centering
    \includegraphics[width=\columnwidth]{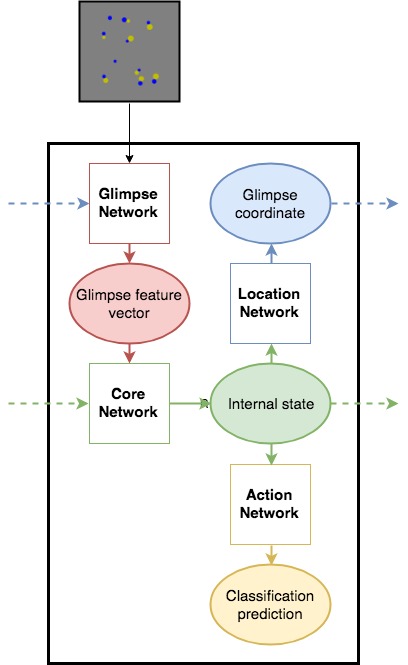}
    \caption{One time-step of the RAM model.}
    \label{fig:ram}
\end{figure}
\begin{itemize}
    \item The glimpse network. It takes the environment (which in the current experiment is the image stimulus) and a location co-ordinate as its inputs. At $t_0$, the location co-ordinate is randomly generated. At all subsequent $t$s, it is selected by the location network (described below) at $t-1$. The network takes a series of samples centred around the co-ordinate and concatenates them into a glimpse. Each consecutive sample is larger than the previous, but at a lower resolution. We used 2 samples, the second of which was twice as large and at half the resolution of the first. These are then processed by 3 convolutional layers and one fully-connected ReLU layer to generate a ``what'' vector. In parallel, the co-ordinate is processed by a ReLU layer outputting a ``where'' vector. The ``what'' and ``where'' vectors are point-wise multiplied to generate the glimpse feature vector.
    \item The core network. An LSTM cell, which takes the glimpse feature vector at $t$ and its own internal state at the previous time-step as its inputs. 
    \item The location network. A fully connected layer which takes the core network's internal state at $t$ as its input, and outputs two values ranging between $-1$ and $1$ (via tanh) as its output.  These are the means of Gaussians (we fixed the standard deviation at 0.03), one for the $x$ coordinate and one for the $y$.  Actual coordinates are samples from them and are fed in to the glimpse network at $t+1$.
    \item The action network. A fully connected layer which takes the core network's internal state at $t$ as its input and outputs a binary image classification. The action network produces a classification at every $t$, but we only record the classification decision that occurs at the final $t$.
\end{itemize}

Neural networks are not bound by `wall clock time', so it is not possible to directly manipulate the amount of time they use to do a task. To operationalise trial duration, we use the networks' architectures to implement processing constraints which reflect those faced by human subjects operating under urgency.
The operationalisations reflect two complementary ideas about the effect that task duration will have on human speakers: as duration increases, (i) the amount of information processing and (ii) the number of saccades and fixations possible increases. For CNNs, we manipulate network depth (thus manipulating the amount of information processing possible) and for the RAM model, we manipulate the number of glimpses made by the network. Each experiment used four levels of task duration: we use the VGG7, 9, 11 and 13 architectures and RAM networks with 4, 8, 16 and 24 glimpses. 

The VGG models are trained using the Adam optimizer \cite{Kingma2015}. For the RAM models, we adopted the hybrid supervised learning approach described in \citet{Mnih2014}, where cross-entropy is back-propagated to train the action, core, and glimpse networks, and the REINFORCE rule \cite{Williams1992, Sutton1999} is used for the location network.  Complete hyper-parameters and training details are included in the Supplementary Materials section.
The source code and data may be found at \url{https://github.com/shanest/neural-vision-most}.

\section{Results}

\subsection{Descriptive}  
\begin{figure}[ht]
    \centering
    \includegraphics[width=\columnwidth]{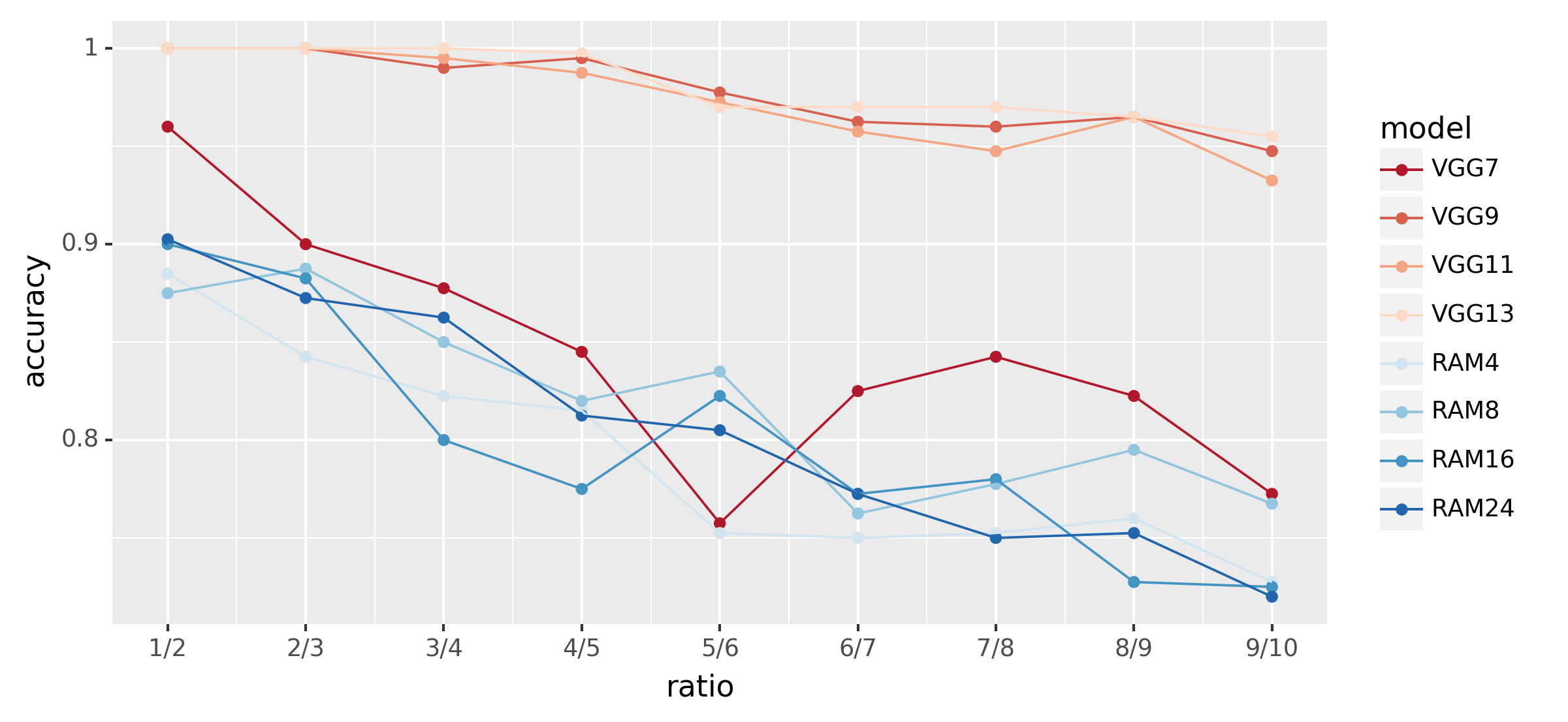}
    \caption{Accuracy by ratio, across image type.}
    \label{fig:by_ratio}
\end{figure}

Figure~\ref{fig:by_ratio} shows the accuracy of all networks by dot ratio, averaged across all image types. In both network types, there is a clear trend of decreasing accuracy as ratios become more balanced. There is a notable clustering of the three VGG9+ networks: they appear to have very similar accuracies across all ratios, and follow a pattern that differs dramatically from VGG7, which is significantly less accurate. Notably, VGG7 is the only CNN network not to attain 100\% accuracy at any ratio.  Moreover, its performance collapses much more rapidly than the other CNNs as the ratios become more balanced. Whilst the RAM networks appear to cluster together a bit, their performance at each ratio shows a broader degree of variability at each ratio than the CNNs.

\begin{figure}[hb]
    \centering
    \includegraphics[width=\columnwidth]{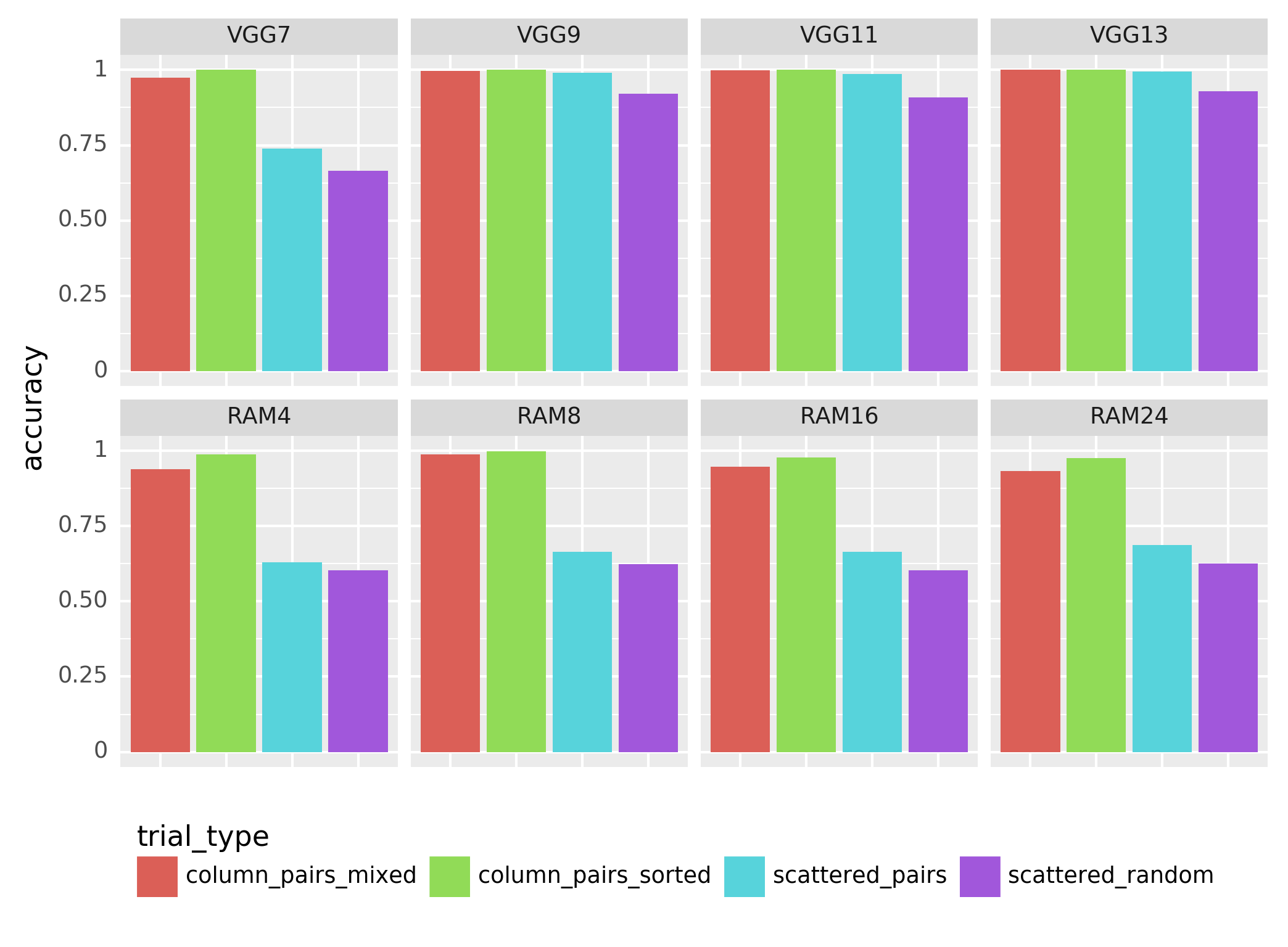}
    \caption{Accuracy by trial type, averaged across ratio.}
    \label{fig:by_trial}
\end{figure}

Figure~\ref{fig:by_trial} shows the accuracy of all networks by image type, averaged across all ratios. As above, there are easily observable differences in the performance of the VGG7 and VGG9+ networks. The former performed more poorly on the scattered type images than the column types, whereas the latter attained near-or-at-ceiling accuracy on all but the scattered random trials. The RAM networks' response pattern was similar to VGG7's, albeit somewhat more pronounced. With the exception of instances where near-or-at-ceiling responses make the data less legible, for the column and scattered image sets, all networks performed more accurately on the image types that contained paired dots than their unpaired equivalent.

\begin{figure*}[ht]
    \centering
    \begin{tabular}{cc}
    \includegraphics[width=0.45\textwidth]{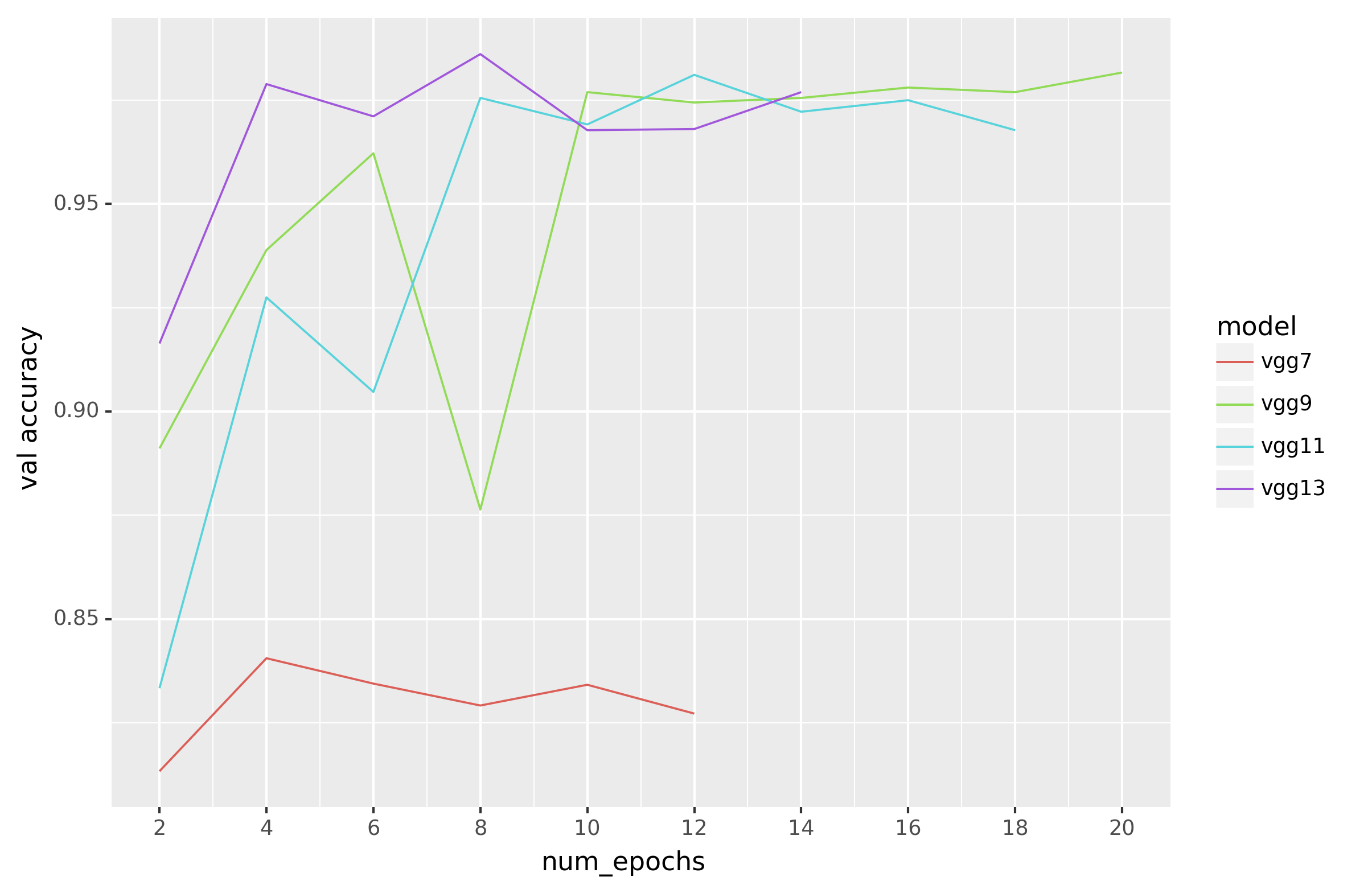}
    &
    \includegraphics[width=0.45\textwidth]{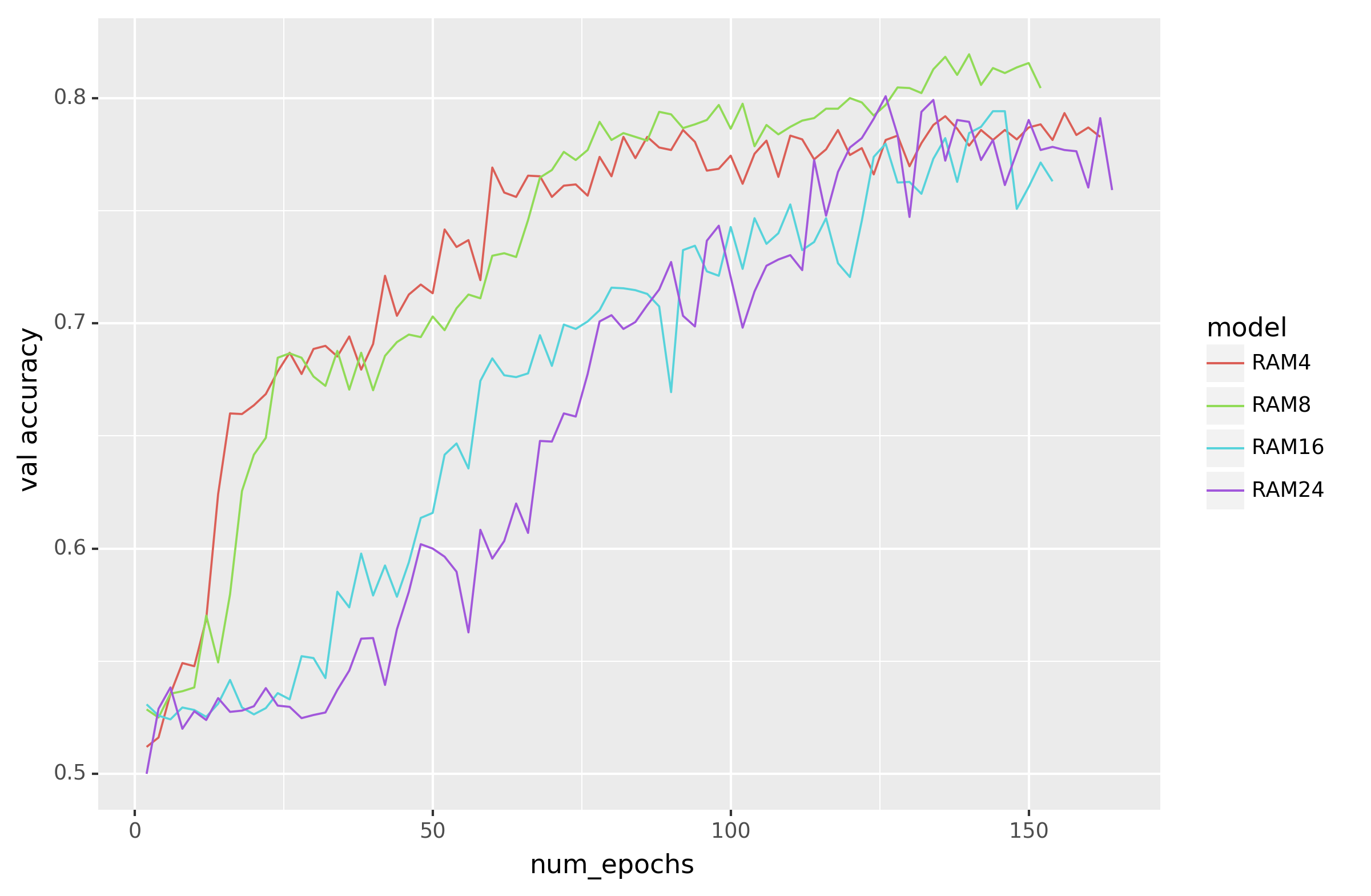}
    \end{tabular}
    \caption{Learning curves for VGG (left) and RAM (right) models.}
    \label{fig:learning_curves}
\end{figure*}

Figure~\ref{fig:learning_curves} shows the learning curves for both model types.  The VGG7 model hits peak performance quickly, and does not improve thereafter.  The VGG13 hits near-ceiling performance very quickly.  VGG9 and 11 show more involved learning patterns, with significant decreases in accuracy before hitting their ceilings.  These results reflect the VGG7's limited computational capacity relative to the other three.

The RAM models show an interesting pattern: the two models with fewer glimpses (4 and 8) have very similar learning trajectories, as do the two models with more glimpses (16 and 24).  And while all four end up at roughly the same accuracy, the former models begin improving much earlier.  This suggests that learning how to choose a large number of glimpse location choices is a difficult reinforcement problem.  A more detailed analysis of model behavior throughout learning will be left for future work.

\subsection{Regression Analysis}  To test the significance of these apparent trends, we fit separate multiple logistic regression models to the data from each network type.\footnote{See, among others, \citet{Kotek2011} (\S 3.3.2) for a multiple logistic regression analysis of experimental data on truth-value judgments of `most' sentences.} Correct prediction was the outcome variable. Three predictor variables relating to the hypotheses were included: image type, a categorical variable; operationalised task duration, an ordinal variable; and dot ratio (converted to real numbers), a continuous variable. Dot ratio was ordered from least balanced (1/2) to most balanced (9/10).  We also included two control predictor variables to verify whether dot ratio is the primary explanatory variable for differences in performance following manipulations of dot ratio sizes, rather than related or potentially confounding factors. These were: absolute set size difference, a continuous variable; and total dots, a continuous variable. The model also included one interaction term, between dot ratio and network type. The CNN model could not produce reliable statistical estimates for some variable levels due to response invariance (i.e. when performance was at-or-near-ceiling, there were not enough incorrect predictions to reliably estimate paramters). As such, these were excluded from the analysis. These were the data corresponding to column-type images and VGG13. Of the variables included in each analysis, the network with the greatest operationalised task duration (i.e. VGG11 and RAM24) and the `most organised' image type (i.e. scattered pairs in the CNN analysis and column pairs sorted in the RAM analysis) acted as the comparison class.

\begin{table}[ht]
\centering
\begin{adjustbox}{max width=\columnwidth}
\begin{tabular}{lllll}
\toprule
Variable & Estimate                                & Std. Error                                                     & $z$ value & $Pr(>|z|)$                  \\
\midrule
Image: Scattered pairs (Intercept) & 16,20                                                          & 4,15    & 3,91                  & 9.42e-05 ***        \\
Image: Scattered random            & -0,79                                                          & 0,09    & -8,63                 & \textless 2e-16 *** \\
Network: VGG9                     & -0,73                                                          & 3,47    & -0,21                 & 0,83                \\
Network: VGG7                     & -12,22                                                         & 2,45    & -4,98                 & 6.37e-07 ***        \\
Dot ratio                               & -14,90                                                         & 4,93    & -3,02                 & 0.00253 **          \\
Absolute difference                     & 0,17                                                           & 0,37    & 0,46                  & 0,64                \\
Total dots                              & -0,03                                                          & 0,04    & -0,87                 & 0,39                \\
Ratio * Network: VGG9          & 1,09                                                           & 4,00    & 0,27                  & 0,78                \\
Ratio * Network: VGG7          & 11,81                                                          & 2,83    & 4,18                  & 2.97e-05 ***        \\
\bottomrule
\end{tabular}
\end{adjustbox}
\caption{Multiple logistic regression of the CNN trials.  Significance:  0 `***' 0.001 `**' 0.01 `*' 0.05 `.' 0.1.}
\label{tab:cnnregression}
\end{table}

The output of the CNN logistic regression can be seen in Table~\ref{tab:cnnregression}. The model shows that the log-odds of the VGG7-11 networks correctly predicting a stimulus' label are significantly reduced as the stimulus' dot set ratio becomes more balanced. We found no significant effect for either of our control variables (absolute difference and total number). These findings strongly support Hypothesis 1. Holding all other variables constant, VGG7-11 were significantly less likely to predict the correct label of scattered random images than scattered pairs images. Given that the lack of difference between the images types that could not be included in the analysis appears to be due to ceiling effects, we interpret these findings as supporting Hypothesis 2. Holding all other variables constant, VGG7 was significantly less likely than VGG11 to make a correct classification. No difference was found between VGG9 and 11. Again, as the lack of difference between the VGG9+ networks appears to be best explained by response invariance due to ceiling effects, we cautiously interpret these findings as supporting Hypothesis 3. Finally, we found a significant positive interaction between dot ratio and VGG7.  Together with the negative coefficient for VGG7, the result is that the predicted log-odds for a correct prediction by VGG7 are robustly lower across ratios than for VGG9 and VGG11, as expected.  The positive interaction term means that the log-odds decrease at a slower rate for more balanced ratios for VGG7 than the other two; this is due to the at-or-near-ceiling performance of the other two at many of the less-balanced ratios.

\begin{table}[ht]
\centering
\begin{adjustbox}{max width=\columnwidth}
\begin{tabular}{lllll}
\toprule
Variable & Estimate & Std. Error & $z$ value & $Pr(>|z|)$ \\
\midrule
Image: Column pairs sorted (Intercept) & 9,57     & 1,52       & 6,28    & 3.41e-10 ***          \\
Image: Column pairs mixed              & -1,18    & 0,16       & -7,55   & 4.37e-14 ***          \\
Image: Scattered pairs                 & -3,54    & 0,14       & -25,15  & $<$ 2e-16 ***    \\
Image: Scattered random                & -3,75    & 0,14       & -26,73  & $<$ 2e-16 ***   \\
Glimpses: RAM16                       & -0,32    & 0,51       & -0,63   & 0,53                  \\
Glimpses: RAM8                        & -0,97    & 0,51       & -1,91   & 0,06                  \\
Glimpses: RAM4                        & -0,77    & 0,50       & -1,54   & 0,12                  \\
Dot ratio                                   & -6,91    & 1,84       & -3,75   & 0.000179 ***          \\
Absolute difference                         & -0,25    & 0,15       & -1,64   & 0,10                  \\
Total dots                                  & 0,04     & 0,02       & 2,24    & 0.025427 *            \\
Ratio * Glimpses: RAM16           & 0,33     & 0,63       & 0,52    & 0,60                  \\
Ratio * Glimpses: RAM8            & 1,34     & 0,62       & 2,14    & 0.032372 *            \\
Ratio * Glimpses: RAM4            & 0,81     & 0,62       & 1,31    & 0,19                  \\
\bottomrule
\end{tabular}
\end{adjustbox}
\caption{Multiple logistic regression on RAM trials.}
\label{tab:ramregression}
\end{table}

The output of the RAM logistic regression can be seen in Table~\ref{tab:ramregression}. According to the model, the log odds of a RAM network correctly labelling stimuli is significantly reduced as set ratios become more balanced. We also found a small but significant effect of total dots, indicating that the likelihood of a correct prediction increases with total dots. This is unsurprising, as increasing total dots reduces image sparseness, increasing the odds that glimpses will contain dots.  This can be especially important for the initial glimpse, which has a random location. This does not invalidate the dot ratio finding, given their comparative effect sizes. No significant effect was found for absolute difference. These findings support Hypothesis 1. The log odds of a RAM network predicting the correct labels for column pairs mixed, scattered pairs or scattered random images was significantly lower (by varying degrees) than for column sorted pairs images. This strongly supports Hypothesis 2. We found no significant difference in the likelihood of the 4-16 glimpse RAM networks correctly labelling stimuli than their comparison class, the 24 glimpse RAM network. These findings do not support Hypothesis 3. Finally, we found a small but significant positive interaction between dot ratio and RAM8, suggesting that the increase in log-odds of correct prediction per unit increase in dot ratio is stronger for RAM8 than for RAM24.  Because the effect size is small, we caution against over-interpreting this result.  And, as before, this effect is somewhat offset by a negative coefficient for RAM8, lowering the intercept in this case.

\subsection{ANS Model Fitting}  
\begin{figure}[ht]
    \centering
    \begin{tabular}{c}
    \includegraphics[width=\columnwidth]{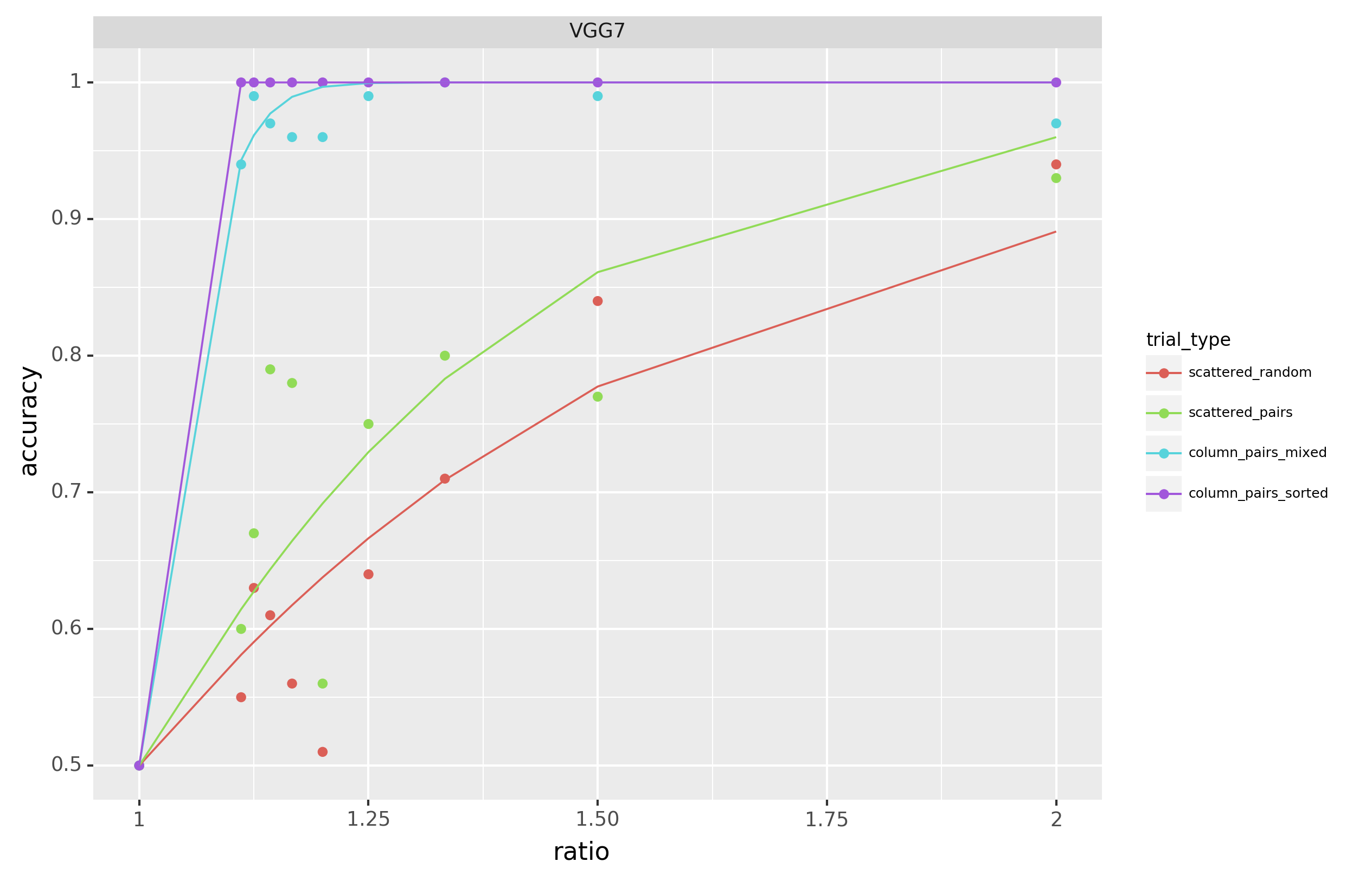} 
    \\
    \includegraphics[width=\columnwidth]{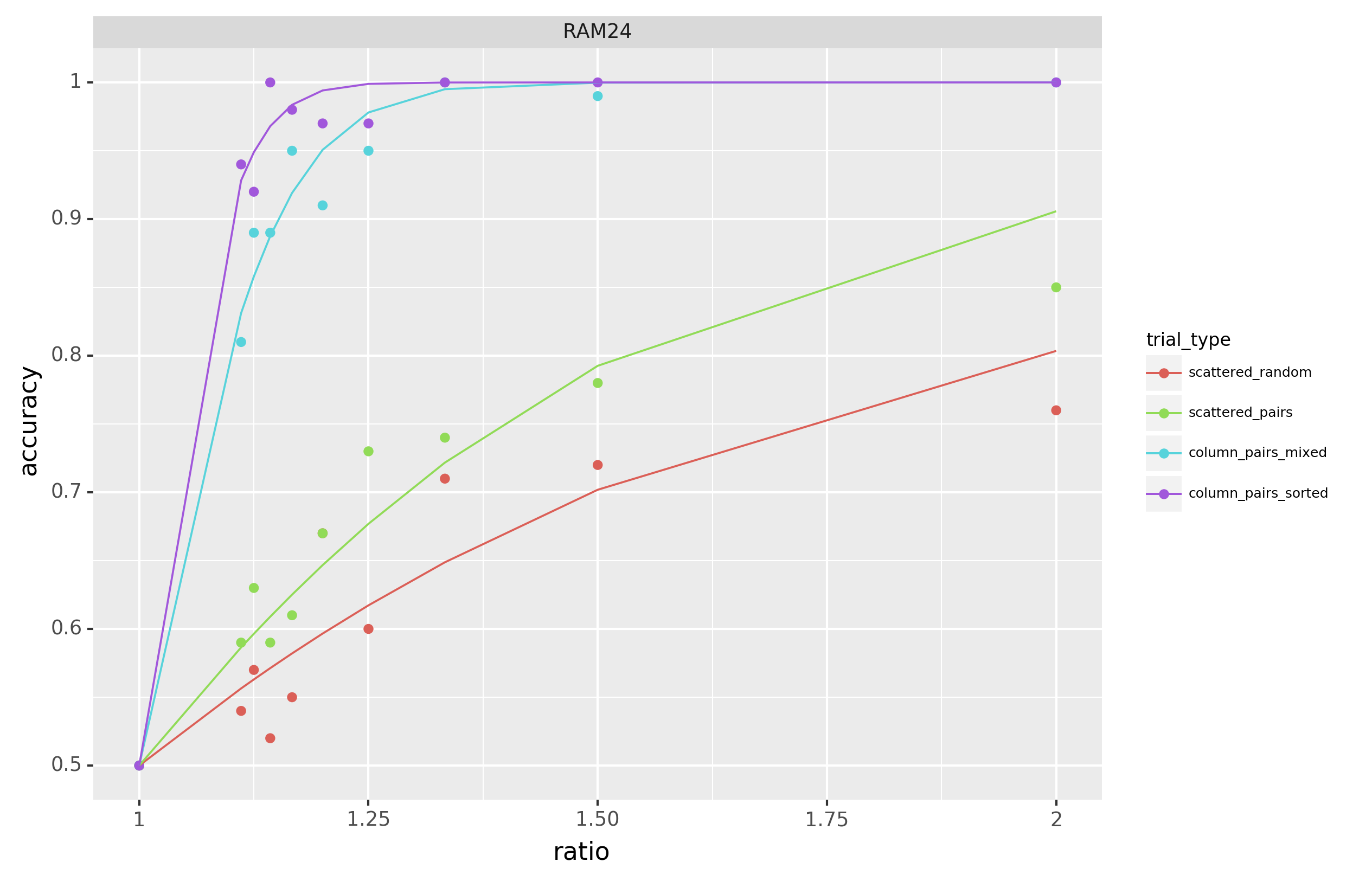}
    \end{tabular}
    \caption{Accuracy by trial type and ratio for VGG7 (top) and RAM24 (bottom), with Weber curves.}
    \label{fig:weber}
\end{figure}

For each model, we also fit a psychophysical model of the Approximate Number System (ANS) to the mean accuracy data, broken down by ratio and by image type \cite{Pica2004, Nieder2004, Halberda2008}.  For this model, ratio is ordered from most balanced (10/9) to least balanced (2/1).  The model represents numerosities as Gaussians, and comparisons between numerosities via the difference in Gaussians.  In particular, there is one free parameter $w$ -- the Weber fraction -- which represents increase in accuracy with increase in ratio.  More precisely, we fit the following model:
$$\text{acc} = 1 - \frac{1}{2}\text{erfc}\left(\frac{n_1 - n_2}{w\sqrt{2}\sqrt{n_1^2 + n_2^2}}\right)$$
where $n_1$ represents the larger number and $n_2$ the smaller.
Figure~\ref{fig:weber} shows the fit curves for the VGG7 and RAM24 networks, which exhibited the most human-like behavior.  An Appendix includes these for all eight models.

For both networks, both column image types have a significantly higher degree of accuracy than both scattered types, with scattered pairs being a bit easier than scattered random.  The psychophysical model provides a good fit to the data: Table~\ref{tab:weber} provides the Weber fractions and $R^2$ for these cases. 
For human participants, \citet{Pietroski2009} found $w$ to be roughly $0.3$ on all but the column sorted trials, where $w$ was $0.04$.  Our models are not too far off of these Weber fractions, with one noticeable difference: our models treat column mixed trials much more similarly to column sorted trials, whereas for humans column mixed trials pattern with the two scattered trial types.   

\begin{table}[hbt]
    \centering
    \begin{adjustbox}{max width=\columnwidth}
    \begin{tabular}{ccccc}
    \toprule
    & \multicolumn{2}{c}{VGG7} & \multicolumn{2}{c}{RAM24} \\
    type & $w$ & $R^2$ & $w$ & $R^2$ \\
    \midrule
    scattered random & 0.363 & 0.843 & 0.524 & 0.801 \\
    scattered pairs & 0.256 & 0.581 & 0.340 & 0.913 \\
    column mixed & 0.047 & 0.979 & 0.078 & 0.975 \\
    column sorted & 0.012 & 1.0 & 0.051 & 0.984 
    \\
    \bottomrule
    \end{tabular}
    \end{adjustbox}
    \caption{Weber fractions and $R^2$ for the ANS model.}
    \label{tab:weber}
\end{table}

\section{Discussion}

We subjected convolutional networks of varying depths and recurrent models of visual attention with varying number of glimpses to the psychosemantic experiment of \citet{Pietroski2009}.  Our first two hypotheses are confirmed: all networks show decreased accuracy with decreasing dot ratio as well as a strong sensitivity to image type.  The third hypothesis is partially confirmed: increasingly deep CNNs do show increased performance (with all VGG9+ networks being near ceiling), while increasing the number of glimpses for a RAM model has little effect on overall accuracy.  The psychophysical model of approximate number fits network data well, with some Weber fractions being near those found for human participants.  For the RAM models, this suggests that visual attention and search may be a causal mechanism underlying some ANS-like responses.  The primary qualitative difference between model performance and human performance is that the models do roughly equally well on both column image types, whereas humans are significantly better on column sorted as opposed to column mixed trials.  This suggests that the strategies learned by the models differ in some interesting ways from those employed by human participants.

These results exhibit initial promise in using neural models as cognitive models in psychosemantics.  In particular, while the fit with existing human data is good (criterion (i) above), it is not quite strong enough to warrant generating robust predictions about manipulations like task duration (criterion (ii) above).  Nevertheless, these initially promising results also suggest interesting avenues for future work.  

(1) More detailed hyper-parameter searches may improve fit with the human data, thus allowing us to use the models to generate predictions.  (2) RAM model performance could be improved by giving the network a low-resolution version of the whole image to help it make location choices \cite{Ba2014}. (3) While our depth manipulation for CNNs was designed to reflect increased information processing capacity as duration increases, one could control for capacity (number of parameters in the model) by making the deeper networks narrower or the shallow networks wider, and seeing if depth still has an effect. (4) To better understand what strategies the models are using to solve the task, techniques such as transfer learning and diagnostic classifiers \cite{Hupkes2018, Giulianelli2018} could be applied to our models.  (5)  Similarly, one can investigate whether any neurons or groups thereof in the models exhibit activation curves consistent with Weber's law \cite{Nieder2004}. (6) Independent neural models that exhibit ANS-like behavior---or, more generally, that are trained on other image processing tasks---could be used in this task \cite{Stoianov2012}; a key challenge here will be operationalizing task duration.  (7) The models could be used to model performance against more image manipulations, such as the number of colors in a scence \cite{Lidz2011}.  We leave these and other avenues for improving neural models of psychosemantic tasks to future work.

\section*{Acknowledgments}

Thanks to Alexandre Cremers and Jakub Szymanik for helpful discussion and to three anonymous CMCL reviewers for valuable comments.  We thank SURFsara (\url{www.surfsara.nl}) for the support in using the Lisa Compute Cluster. SS-T has received funding from the European Research Council under the European Union's Seventh Framework Programme (FP/2007–2013)/ERC Grant Agreement n. STG 716230 CoSaQ.

\bibliography{Papers-neural-most}
\bibliographystyle{acl_natbib}

\appendix

\section{Supplementary Material}

Images were 128x128 pixels, converted to grayscale.  The TensorFlow Python library \cite{Abadi2016} was used to implement everything. The networks were trained and tested on an NVIDIA GeForce 1080Ti GPU.  
The source code and data may be found at \url{https://github.com/shanest/neural-vision-most}.

The RAM models had the following hyper-parameters (found by a small grid search):
\begin{itemize}
    \item Number of patches: 2
    \item Size of patches: 12, 24 pixels
    \item Glimpse network: 
    \begin{itemize}
        \item three convolutional layers with 64, 64, and 128 filters and kernel size 5, 3, and 3, respectively
        \item Output vector size: 512
    \end{itemize}
    \item Core network: LSTM with hidden state dimension 1024
\end{itemize}
We trained using the Adam optimizer with learning rate 1e-5.  The RAM models were trained for up to 200 epochs, with early stopping with a patience of 10 epochs (i.e.\ training was stopped when loss did not improve over a ten epoch time-frame, as measured every 2 epochs).

The CNN models were trained using 0.25 dropout (on the final fully-connected layers) and the Adam optimizer, with learning rate 1e-4.  We used early stopping with a patience of 10 epochs, with maximum training length of 40 epochs.  For each model, we saved the best version, as measured by loss on the validation set.

\section{Appendix}

\begin{figure}[ht]
    \centering
    \includegraphics[width=\columnwidth]{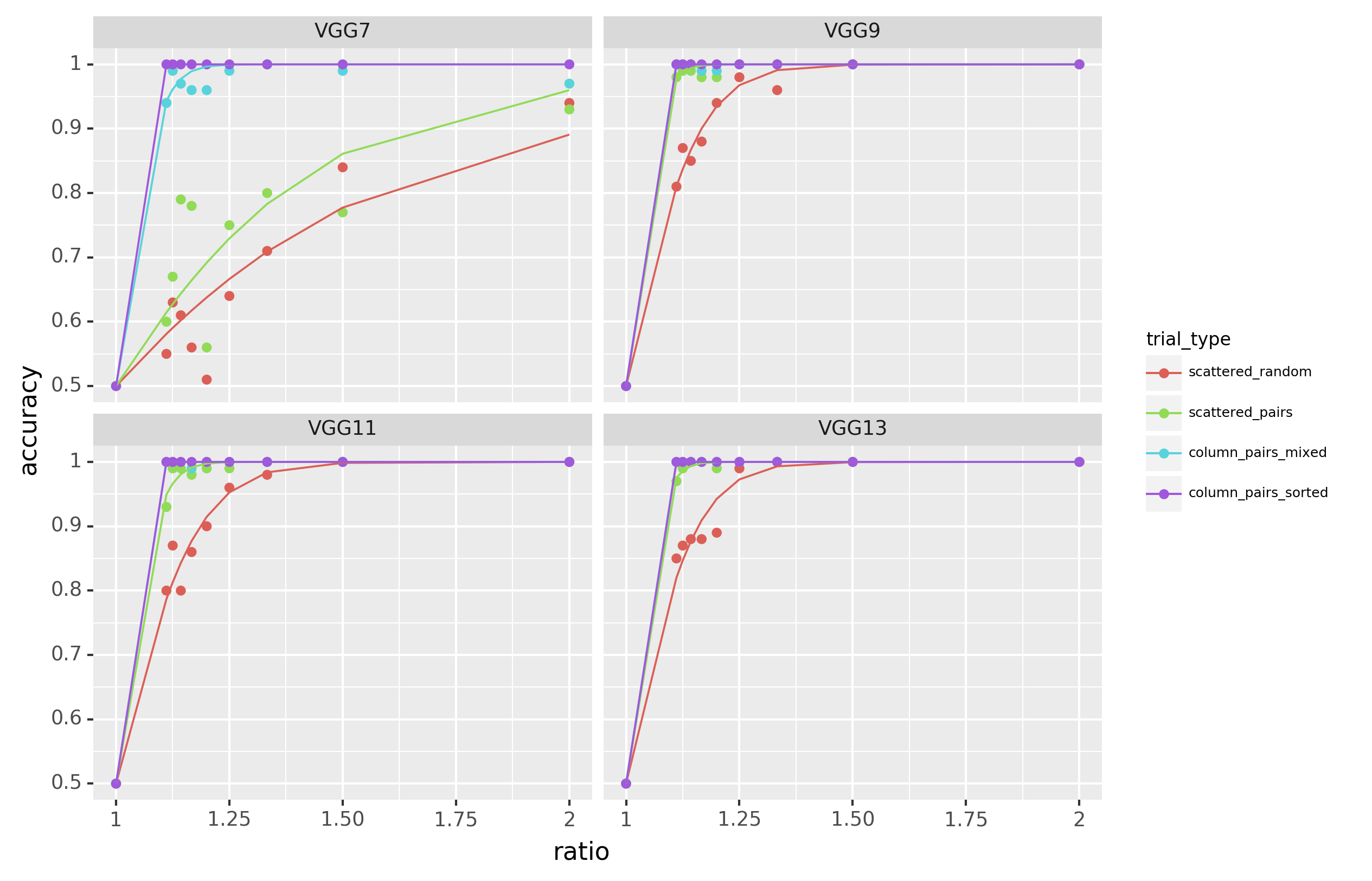}
    \caption{Fit Weber curves for all VGG models.}
    \label{fig:vgg_fit}
\end{figure}

\begin{figure}[ht]
    \centering
    \includegraphics[width=\columnwidth]{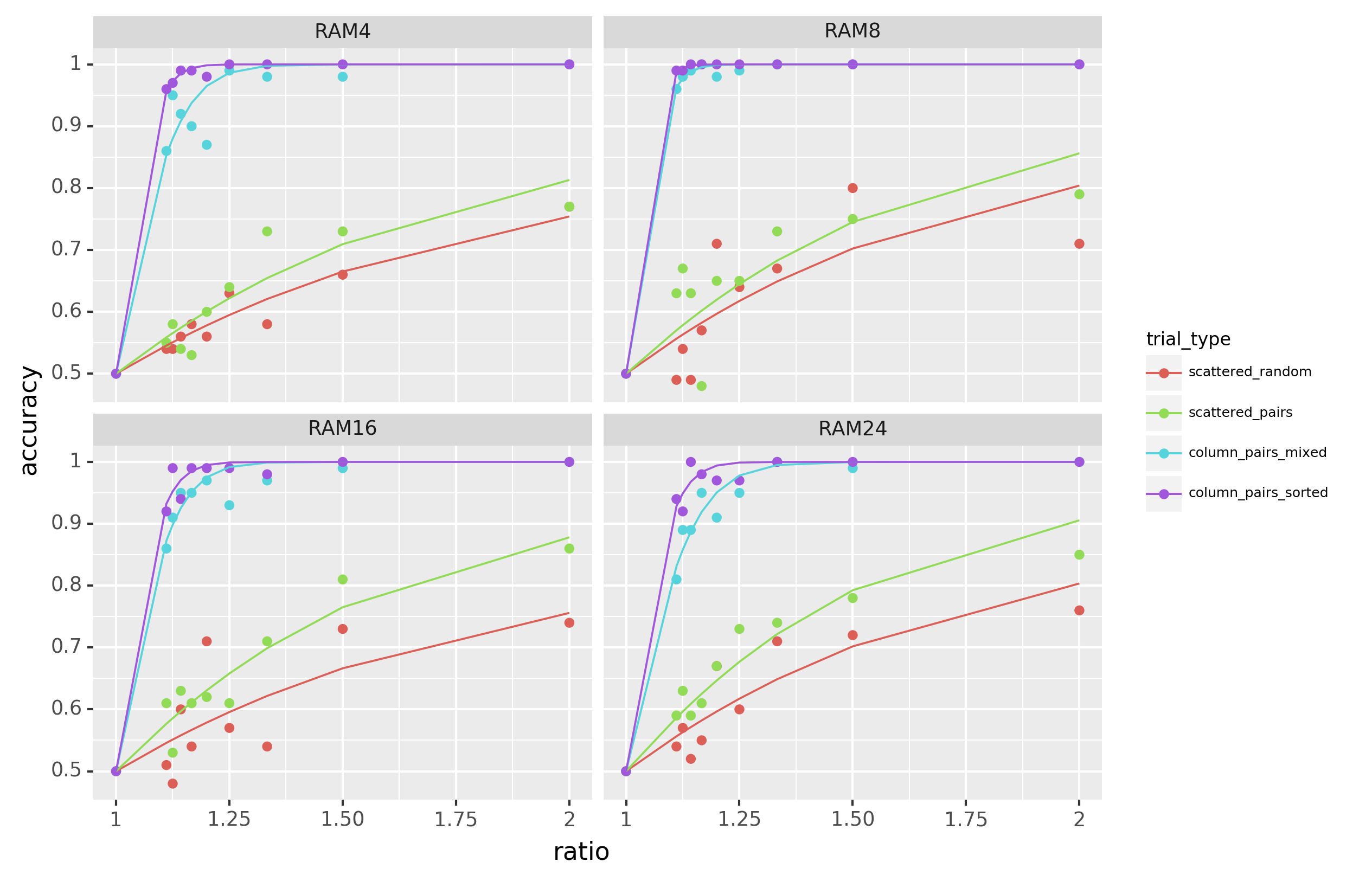}
    \caption{Fit Weber curves for all RAM models.}
    \label{fig:ram_fit}
\end{figure}

Here we include results of fitting the psychophysical model of approximate number to all 8 of our models. Figure~\ref{fig:vgg_fit} shows the VGG models, and Figure~\ref{fig:ram_fit} shows the RAM models.

As can be seen, VGG9-13 look very similar, with the only non-ceiling performance coming on scattered random trials, which it still learns perfectly for large enough (imbalanced enough ratios).  VGG7 shows highly ratio-dependent performance for both scattered random and scattered pairs trials.

The RAM models show subtle patterns.  The model with 8 glimpses (RAM8) was very slightly the best overall performer, but this looks due to the two column trial types.  RAM24 appears to have the best performance on the scattered trial types, at the expense of the column types.  For the scattered types, performance is approaching human levels of accuracy (roughly 89\%, compared to the model being roughly 85\%).

As mentioned in the paper, all models perform similarly on both column trial types, in contrast to human participants, who are significantly better on column sorted than column mixed trials.

Table~\ref{tab:weber_fits_all} provides the results of fitting the psychophysical model to mean accuracy for each model and trial type.  In particular, we report the one parameter of the model (Weber fraction, $w$), and the goodness of fit of each model ($R^2$).

\begin{table*}[ht]
    \centering
    \begin{tabular}{ccccccccc}
    \toprule
    & \multicolumn{2}{c}{scattered random}
    & \multicolumn{2}{c}{scattered pairs}
    & \multicolumn{2}{c}{column mixed}
    & \multicolumn{2}{c}{column sorted}
    \\
    model 
    & $w$ & $R^2$
    & $w$ & $R^2$
    & $w$ & $R^2$
    & $w$ & $R^2$ \\
    \midrule
    VGG7 & 0.363 & 0.843 & 0.256 & 0.581 & 0.047 & 0.978 & 0.012 & 1.0 \\
    VGG9 & 0.085 & 0.985 & 0.085 & 0.997 & 0.015 & 0.999 & 0.012 & 1.0 \\
    VGG11 & 0.093 & 0.971 & 0.045 & 0.994 & 0.015 & 0.999 & 0.012 & 1.0 \\
    VGG13 & 0.081 & 0.973 & 0.038 & 0.999 & 0.012 & 1.0 & 0.012 & 1.0 \\
    RAM4 & 0.0650 & 0.929 & 0.503 & 0.845 & 0.071 & 0.917 & 0.043 & 0.998 \\
    RAM8 & 0.522 & 0.593 & 0.420 & 0.592 & 0.042 & 0.998 & 0.033 & 0.999 \\
    RAM16 & 0.646 & 0.574 & 0.384 & 0.912 & 0.049 & 0.986 & 0.049 & 0.986 \\
    RAM24 & 0.524 & 0.801 & 0.340 & 0.913 & 0.078 & 0.975 & 0.051 & 0.984
    \\
    \bottomrule
    \end{tabular}
    \caption{Weber fractions ($w$) and correlations ($R^2$) for all models and all trial types.}
    \label{tab:weber_fits_all}
\end{table*}

\end{document}